\documentclass[conference,compsoc]{IEEEtran}
\IEEEoverridecommandlockouts
\usepackage{cite}

\usepackage{amsmath,amssymb,amsfonts}
\usepackage{graphicx}
\usepackage{textcomp}
\usepackage{algorithm2e}
\usepackage{float}

\usepackage{xcolor}
\usepackage{subcaption}
\def\BibTeX{{\rm B\kern-.05em{\sc i\kern-.025em b}\kern-.08em
    T\kern-.1667em\lower.7ex\hbox{E}\kern-.125emX}}

\begin{document}

\title{Exploring  Machine Learning Privacy/Utility trade-off from a hyperparameters Lens}



\author{Ayoub Arous$^{1}$, Amira Guesmi$^{1}$, Muhammad Abdullah Hanif$^{1}$, Ihsen Alouani$^{2}$, and Muhammad Shafique$^{1}$
\thanks{$^{1}$Ayoub Arous, Amira Guesmi, Muhammad Abdullah Hanif, and Muhammad Shafique are with eBrain Laboratory, NYU Abu Dhabi, UAE
        }%
\thanks{$^{2}$Ihsen Alouani is with CSIT, Queen’s University Belfast, UK
        }%
}
\maketitle
\newcommand{\ihsen}[1]{{\textcolor{red}#1}{Ihsen: #1}}

\begin{abstract}
Machine Learning (ML) architectures have been applied to several applications that involve sensitive data, where a guarantee of users' data privacy is required. Differentially Private Stochastic Gradient Descent (DPSGD) is the state-of-the-art method to train privacy-preserving models. However, DPSGD comes at a considerable accuracy loss leading to sub-optimal privacy/utility trade-offs. Towards investigating new ground for better privacy-utility trade-off, this work questions; (i) if models' hyperparameters have any inherent impact on ML models' privacy-preserving properties, and (ii) if models' hyperparameters have any impact on the privacy/utility trade-off of differentially private models.
We propose a comprehensive design space exploration of different hyperparameters such as the choice of activation functions, the learning rate and the use of batch normalization. Interestingly, we found that utility can be improved by using Bounded RELU as activation functions with the same privacy-preserving characteristics. With a drop-in replacement of the activation function, we achieve new state-of-the-art accuracy on MNIST (96.02\%), FashionMnist (84.76\%), and CIFAR-10 (44.42\%) without any modification of the learning procedure fundamentals of DPSGD. 


\end{abstract}


\section{Introduction}
\begin{figure}[tp]
    \centering
    {\includegraphics[width=\columnwidth]{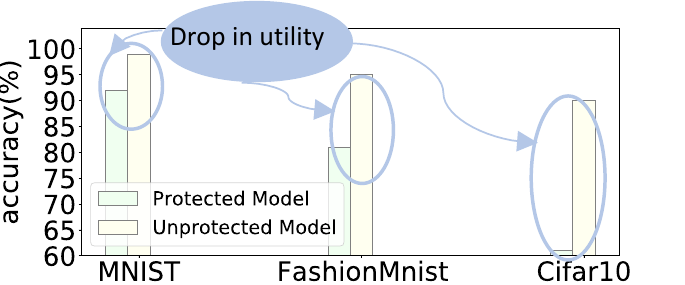}}
    \caption{Comparison of accuracy between protected model with DPSGD and unprotected model for benchmark datasets.}
    \label{intt}
\end{figure}

Deep Learning (DL) models \cite{c1} have gone mainstream in recent years, due to their ability to achieve state-of-the-art performance on a variety of tasks \cite{c2}\cite{aa1}. These models have been used in a variety of applications such as image processing and classification tasks, which require large amounts of data and computational resources to train. However, models also have the potential to compromise the privacy of data used in the training due to this complexity \cite{c3}.
Furthermore, Machine Learning (ML) models are increasingly applied to solving problems that involve sensitive personal information, such as medical records, which leads to serious privacy concerns. Researchers have proposed several attacks on deep learning models such as membership inference attacks \cite{membership}, model stealing attacks \cite{shen} and inversion attacks \cite{he}. Along with these attacks and to address these privacy issues, a plethora of works have proposed new defenses to enhance the models privacy-preserving such as work in
\cite{tr} which focus on providing a framework to detect photo privacy with a convolutional neural network using hierarchical features, and research in \cite{hat} which propose some Adversarial learning techniques for privacy preservation or the use of Generative adversarial networks (GANs) to design models to protect data privacy such as work in \cite{xi}. This race between attacks and defenses has been a challenge and many prior works that provide empirical defenses have been broken by these attacks. However, one of the most important defenses used in the state-of-the-art works and that shows great effectiveness in enhancing the model's privacy is differential privacy \cite{dwork2009differential} which is an emerging defense concept that offers us certain mathematical assurances regarding data privacy, and the most efficient algorithm that incorporates this notion is DPSGD \cite{abadi2016deep}, which is an extension of SGD (see section 2.1 for more details).
Figure \ref{intt} shows the results of DPSGD algorithm on benchmark datasets.  

As Figure \ref{intt} illustrates, there is a major drop in utility for the protected model with DPSGD compared to the unprotected model. So it is obvious that this privacy gain comes with a price of utility loss.
Several related works have tried to find better privacy/utility sweet spots such as work in \cite{bu} which tries to enhance the convergence of DPSGD and research in \cite{shi} which provides a privacy-enhanced matrix factorization with local differential privacy. However, little attention has been directed to the exploration of the potential inherent impact of the hyperparameter choice on the model's privacy-preserving properties. 
To the best of our knowledge, only a few previous studies explored the importance of model hyperparameters in terms of privacy-preserving and resistance to privacy attacks such as MIA. 

\subsection{State-of-the-Art and Their Limitations}
Providing a private model comes with the price of a drop in accuracy. 
So much work has been done to tackle this problem and try to improve the overall trade-off between privacy and accuracy and push the model
to converge to higher accuracy with a lower privacy budget.
To tackle this problem, prior work tried to understand the
cause of a drop in utility with the private model. For
example, \cite{bu2021convergence} provides a theoretical explanation of this drop in accuracy and shows the impact of the added steps in
the differential private model in terms of convergence and
highlight specifically the effect of per simple clipping on
the convergence and try to provide a new variety of clipping
called global clipping based on batch clipping rather than per
simple clipping and show how their method improved this
trade-off between privacy and accuracy.
\par Besides providing new private algorithms some analyses have been conducted to investigate the effects of some parameters on model's outputs which is our case, such as research in \cite{j} and in \cite{juI} which investigates only one activation function, and only one metric is used to evaluate its effectiveness, which is DPSGD. Another work in \cite{bag} investigates the impact of batch size and clipping bound on the model's utility but their analysis has been conducted only on a small model with one dataset (MNIST).
Another research in \cite{papernot2021hyperparameter} shows how a hyperparameter tuning for model parameters could dramatically help DPSGD but this work focused more on the optimization part and provided three optimization algorithms to tune the model and did not focus on providing a wide framework analysis for the model's hyperparameter.
In our scenario compared to work in \cite{j} and in \cite{juI}, we will provide a set of activation functions and contrast them in the contexts of DPSGD and an unprotected model, the impact in the case of an unprotected model will be evaluated using a membership inference attack. Additionally, we offer a theoretical justification for why different activation functions are more effective, more specifically the case of the bounded activation function in DPSGD. By doing this, we are providing a more thorough view of those situations, which will aid more people in trying to implement those concepts. 
Moreover, we will try to conduct our analysis of what is commonly used as datasets and models in a state-of-the-art deep learning analysis. 
\par Finally, the most recent state-of-the-art evaluation of hyperparameters and their effects on model performance lacked a broad framework analysis to clarify those effects and give a more thorough review of the topic.
We are motivated to try to provide a better framework analysis and to try to better explore this topic by those limitations and restricted analysis in this context.

\subsection{Our Novel Contributions} 
Motivated by the importance of investigating new ground for better privacy/utility trade-off, in this paper, we ask the following 2 research questions:
\begin{itemize}
    \item \textbf{Q1 -- } Do models' hyperparameters have an inherent impact on privacy?
    \item \textbf{Q2 -- } If yes, can this be leveraged for better privacy/utility trade-offs in differentially private models?

\end{itemize}

To answer these questions, we propose a comprehensive exploratory study on a set of hyperparameters and investigate their impact on baseline models privacy, as well as their impact on differentially-private models' accuracy under the same privacy budget.

\begin{figure}[tp]
    \centering
    {\includegraphics[width=\columnwidth]{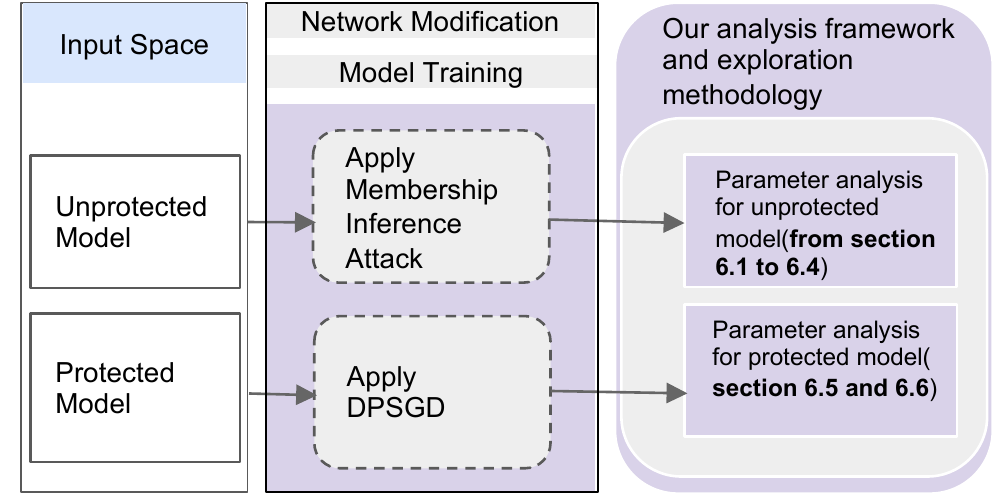}}
    \caption{An overview of our major contributions.}
    \label{intot}
\end{figure}

We discover that several hyperparameters, as well as their combination, assure a greater gain in utility than the typically employed CNN design, thus we present a thorough analysis and demonstrate which parameter is appropriate in which case.
Finally, our contributions will be:
\begin{enumerate}
    \item We use membership inference attack to evaluate the impact of hyperparameter space on attack success rate with deep learning models.
     \item we did a wide analysis in the case of unprotected model, we demonstrate the effect of hyperparameter space on the attack success rate of a membership inference attack and compare the baseline model to the model with the modified parameter.
     \item We did an exploration of the hyperparameter space and its impact in the case of a protected model with DPSGD.
    \item We demonstrate how models with certain hyperparameters could outperform state-of-the-art work when applied with DPSGD.
\end{enumerate}
Finally Figure \ref{intot} summarizes all the major contributions.
\section{Background}
\subsection{Differential Privacy}

\begin{figure}[htp]
    \centering
    {\includegraphics[width=0.7\columnwidth]{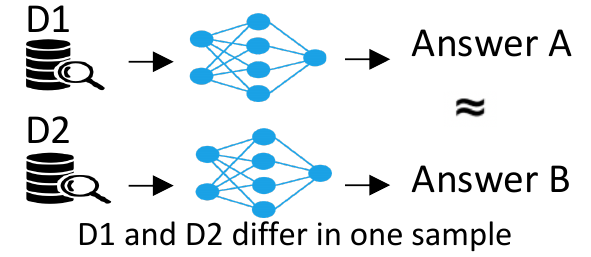}}
    \caption{An overview of Differential Privacy.} 
    \label{inttt}
\end{figure} 

\par A method for protecting privacy when training machine learning models on aggregate datasets is differential privacy highlighted briefly in Figure \ref{inttt}. A survey of the significance of privacy in deep learning can be found in the work \cite{xiong2020comprehensive}. It is defined by the application-specific notion of neighboring databases. For instance, each training dataset used in our study contains a set of image-target pairs. If two of these sets only differ in one item, i.e., if one image-label pair is present in one set but not the other, then we say that the two sets are neighbors. \\
Definition 1. An algorithm  $\mathcal{M}: \mathcal{D} \rightarrow \mathcal{R}$ which run on the input space   $\mathcal{D}$ and range $\mathcal{R}$ is called $(\varepsilon, \delta)$-differential privacy if for any two adjacent datasets $d, d^{\prime} \in \mathcal{D}$ and for any subset of outputs $S \subseteq \mathcal{R}$  it satisfies the following expression:
$$
\operatorname{Pr}[\mathcal{M}(d) \in S] \leq e^\epsilon \operatorname{Pr}\left[\mathcal{M}\left(d^{\prime}\right) \in S\right]+\delta
$$
$\epsilon $ is a measure of privacy that is inversely correlated to the degree of model's privacy. Therefore, with higher epsilon values, protection decreases and user data exposure becomes more likely. DPSGD is a well-known algorithm that uses this idea of differential privacy.
\par Similar to conventional stochastic gradient descent, DPSGD \cite{yagnik2022novel} begins by selecting a batch of data samples at random, computing their gradients, and then updating the models' parameters. The main differences are where Gaussian noise is injected and when the gradient norms are clipped. DPSGD operations include clipping gradients, adding Gaussian noise, and averaging the gradients. As we train, we aggregate the gradients from several mini-batches, and this, too, is differentially private due to the composition properties of differential privacy.
\par The Moments Accountant system then gets to work to better keep track of privacy loss. The moments accountant \cite{van2018three} yields a substantially smaller value of and, hence, a much greater assurance on privacy for a given noise level and number of training steps. The moments accountant, however, allows DPSGD to be run for a certain privacy budget for much more iterations. No matter how we quantify the privacy loss, the learnt model is the same.
\subsection{Membership inference attack}
To determine whether a data sample $x$ was used to train a target model $f_{target}$, membership inference attacks (MIAs) are applied against deep learning models. Consequently, MIA causes a privacy leakage right away, giving the adversaries access to vital training data. For instance, in the real world, $x$ could be a patient's medical file or a person. Because of MIA, the attackers can detect whether another person or a specific patient's clinical file was used to create a model for a certain condition. This clearly violates both privacy and confidentiality.
When the adversary has black-box access to the target model, as is the case in the most common attack scenario, the attackers first train a shadow model $f_{shadow}$ using a shadow dataset $\mathcal{D}_{train}^{shadow}$ , which achieves the same objective as $f_{target}$(e.g., classification). The attackers then use $\mathcal{D}_{train}^{shadow}$(training data) and $\mathcal{D}_{test}^{shadow}$(non-training data) to query $f_{shadow}$, and they successfully retrieve the results. They can create an attack model, in which answers from training data are labeled as 1, and responses from non-training are labeled as 0. Attackers use data sample $x$ to query $f_{target}$ at attack time, then use $f_{attack}$ to determine whether $x$ is used in training the target model or not based on the results from $f_{target}$.
In our exploration this technique is going to be used and we will provide later implementation details, to evaluate privacy leakage via attack success rate which is an indicator in $[0,1]$ that indicates more privacy leakage with higher values.
\subsection{Genetic Algorithm}
\begin{figure}[tp]
    \centering
    {\includegraphics[width=0.7\columnwidth]{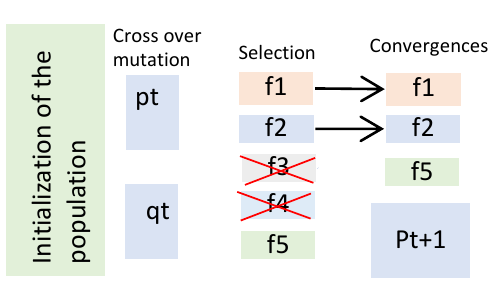}}
    \caption{An overview of  genetic algorithm. A random starting population of chromosomes is generated. Then results for those chromosomes are determined using a performance metric. The best values (with low cost) are subsequently put through reproduction, crossover, and mutation processes. }
    \label{intft}
\end{figure} 

The genetic algorithm\cite{adhikary2022genetic} used in our approach to tune BoundedRELU  added parameter (i.e., the threshold), which is based on natural selection, the mechanism that propels biological evolution, is a technique for resolving both limited and unconstrained optimization issues. At first, chromosomes are generated at random in a population. These parameters values are then submitted to the system model. Results for each set of parameters inside the population are acquired by running a simulation using a performance metric that is based on a cost function. Once each cost value has been determined, it is arranged in ascending order with the corresponding chromosomes. The best values are chosen once more based on which has the lowest cost, and they are then put through processes of reproduction, crossover, and mutation. Figure \ref{intft} shows the basic steps of the genetic algorithm.
 \begin{figure*}[h]
    \centering
    {\includegraphics[width=\textwidth]{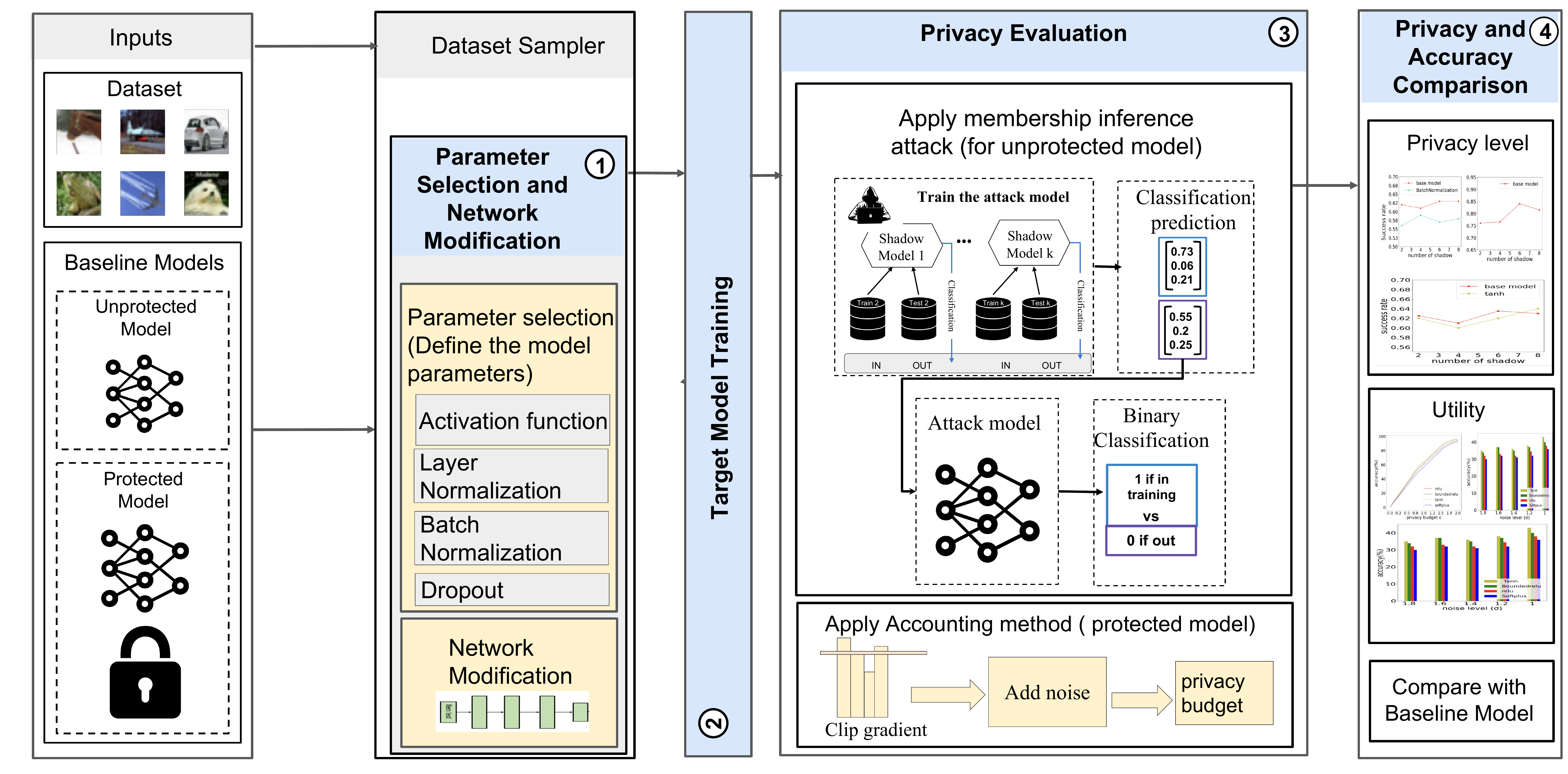}}
    \caption{ An overview of our analysis framework. For each model (protected/unprotected), we start by selecting the parameter to modify our network. Then, we train these modified networks and we apply the privacy evaluation metric (membership inference attack for the unprotected model and DPSGD for the protected model). Finally, we compare the output (privacy/utility) of those modified networks with the output of the baseline models to see the impact of the added parameter.   }
    \label{nn}
\end{figure*}

\section{Analysis framework}
\begin{table}[htp]
\begin{tabular}{ |p{4cm}|p{3.3cm}|  }
\hline
\multicolumn{2}{|c|}{List of Parameters} \\
\hline
Parameter & Values\\
\hline
Learning rate & $\{ e^{-2},e^{-3},e^{-4} \}$  \\
Batch Normalization  & $axis=0$ and $axis=1$\\
Layer Normalization  & $axis=0$ and $axis=1$\\
Dropout & 0.5\\
Activation functions & \{RELU, BoundedRELU, Tanh\}\\

\hline
\end{tabular}
\caption{Hyperparameter space for conducting our analysis.}
\label{table:1}
\end{table}

To conduct our analysis, we use the framework described in Figure \ref{nn}. As the diagram illustrates, our work is divided into two major parts. In the first part, we conduct the analysis on an unprotected model. The structure of the models are mentioned in table \ref{correlations} and \ref{corr}, these models will be referred to as baseline models. We define the model parameters and perform the network modification \textbf{(step1)}, then we train our target model \textbf{(step2)} and we use membership inference attack to evaluate attack success rate on those unprotected models \textbf{(step3)}, finally for each experiment we evaluate the parameter performance by comparing it to the baseline models \textbf{(step4)}. Table \ref{table:1} represents the space of hyperparameters that are used in our analysis. In the second part, we conduct our analysis on the protected model with DPSGD and moment accountant as a tool to assess privacy leakage, and we follow the same methodology mentioned for the unprotected model to assess the impact of hyperparameter space on the trade-off between privacy and utility.

\section{Experimental Methodology}
\begin{figure}[tp]
    \centering
    {\includegraphics[width=1\linewidth]{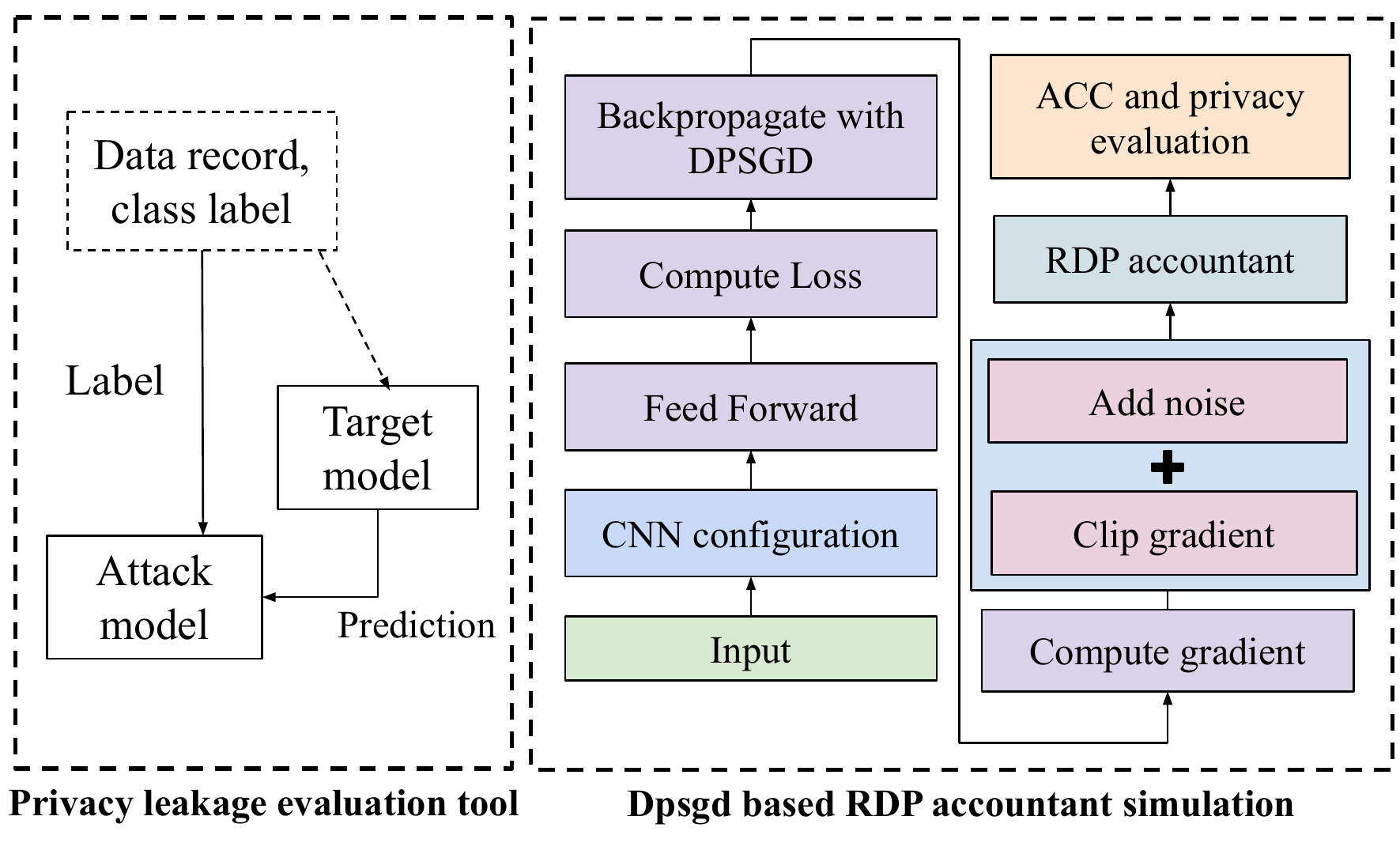}}
    \caption{Experimental setup and tool.}
    \label{e}
\end{figure}

The experiments are conducted on three datasets, MNIST \cite{mnist} that has
60,000 training data and 10,000 test data, The CIFAR-10 \cite{CIFAR} dataset consists of 60000 $32\times32$ color images in 10 classes, with 6000 images per class. There are 50000 training images and 10000 test images and the FashionMnist \cite{fashion} dataset comprising $28\times28$ grayscale images of 70000 fashion products from 10 categories, with 7000 images per category.
In the case of the protected model, we implemented DPSGD for evaluating our protected model using Pytorch \cite{pytorch}. The experiments are performed by training the model for n number of epochs at each noise level and then drawing the privacy budget and accuracy of the last epoch, the noise levels added to the clipped gradient are in the following range: $ list=[1.8,1.6,1.4,1.2,1,0.8] $. The learning rate in the case of DPSGD is fixed to $0.1$ , $delta=1e-5$ and $max-per-sample-grad-norm=1.0$.
In the case of the unprotected model we implemented a membership inference attack using Tensorflow \cite{tensorflow} as follows:\\
The architecture of the target model is shown in Table \ref{corr} for FashionMnist and in Table \ref{correlations} for CIFAR-10.
We then construct the shadow models which inherit the same property of the target model to reveal its pattern to the attack model. We construct the shadow models with the same architecture as the target model for different datasets, and the train and test data of shadow models are sampled from the same target model dataset with the same size. After training the shadow models which we know their ground truth of membership inference attack, we label the itch output vector of shadow models with 0 if it is not in the training data and with 1 if it is in the training data, then we feed those output vectors with their labels to the attack model which is a binary classifier in our case SVM model. After training the attack model we conduct our attack on the target model and report the attack success rate.
The experimental setup for evaluating our methodology is shown in Figure \ref{e}.

For MNIST, we set the threshold of BoundedRELU $a=2$ for a protected model. For FashionMnist, we set $a=3$ and for CIFAR-10 the best performance is for $a=4$. For DPSGD, we use the Opacus package with the RDP accountant method for privacy budget evaluation.\\
For the CNN configuration, the architecture is shown in Table \ref{corr}.
\begin{table}[htp]
\centering
\begin{tabular}{ |c|c|c|c| } 
\hline
Layer & Parameters  \\
\hline
Convolution & 16 filters of 8x8, strides 2\\ 
Max-Pooling& 2x2\\ 
Convolution& 32 filters of 4x4, strides 2\\ 
Max-Pooling& 2x2\\ 
Fully connected & 32 units\\ 
Softmax & 10 units units\\ 
\hline
\end{tabular}
\caption{Convolutional model architecture.}
\label{corr}
\end{table}

\begin{figure}[htp]
    \centering
    {\includegraphics[width=\columnwidth]{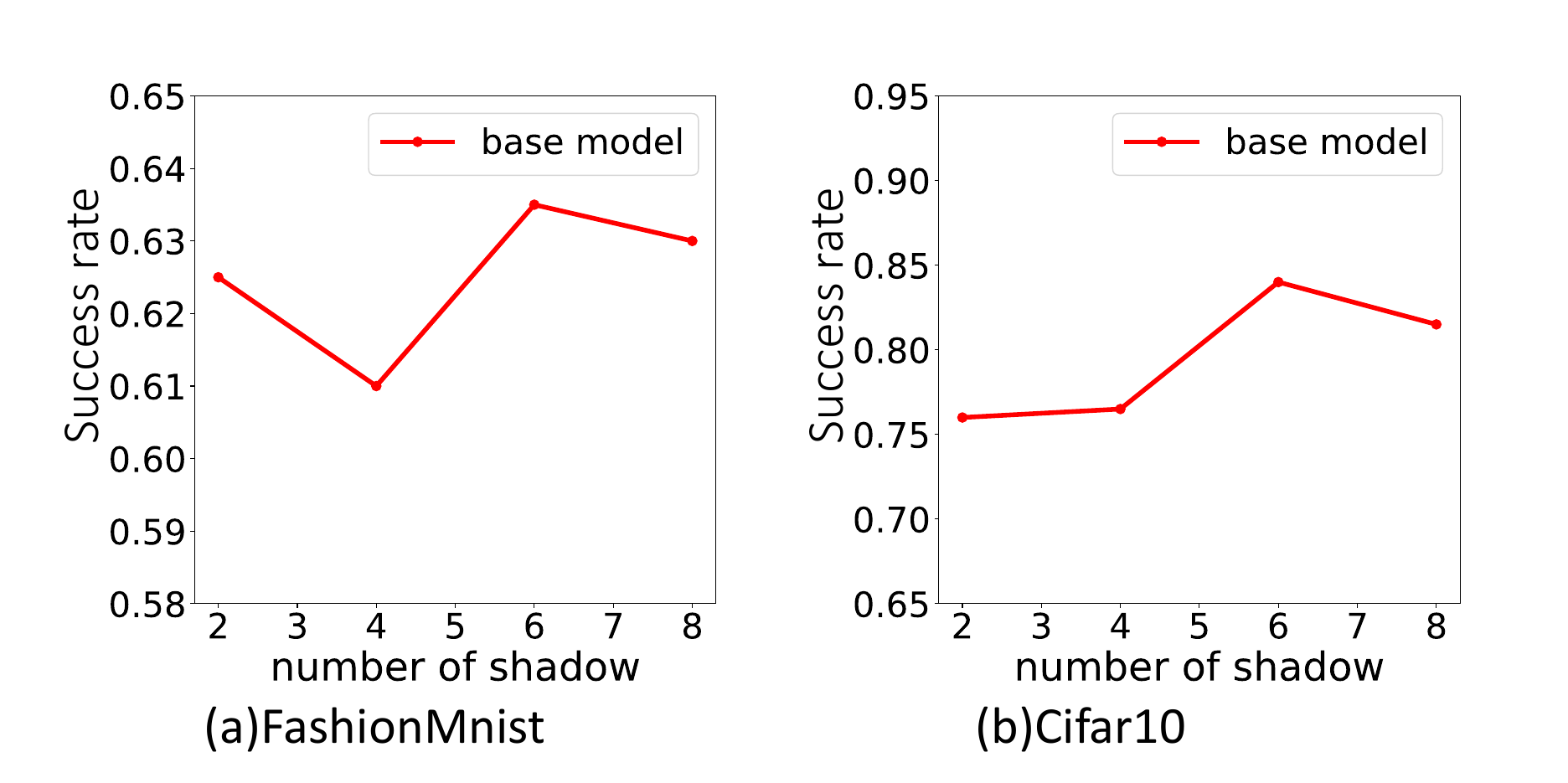}}
    \caption{Results for the baseline models.}
    \label{gk}
\end{figure}

The architecture mentioned in Table \ref{corr} is used for MNIST while the architecture mentioned in Table \ref{correlations} is used to generate the results for CIFAR-10. For the genetic algorithm, the basic block for CIFAR-10 is highlighted in Table \ref{correlations} while we maintain the same architecture in Table \ref{corr} for MNIST dataset. The results for the baseline models are shown in Figure \ref{gk}.
\begin{table}[H]
\centering
\begin{tabular}{ |c|c|c|c| } 
\hline
Layer & Parameters  \\
\hline
Convolution & 32 filters of 3 × 3, strides 1\\ 
Max-Pooling& 2x2\\ 
Convolution& 64 filters of 3 × 3, strides 1\\ 
Max-Pooling& 2x2\\ 
Convolution & 128 filters of 3 × 3, strides 1\\ 
Max-Pooling & 2x2\\ 
Convolution & 256 filters of 3 × 3, strides 1\\ 
Max-Pooling & 2x2\\ 
Fully connected & 32 units\\ 
Softmax & 10 units units\\ 
\hline
\end{tabular}
\caption{CIFAR-10 model architecture for the Genetic Algorithm.}
\label{correlations} 
\end{table}

\section{Experimental Results}
\subsection{Impact of varying activation functions on attack success rate for an unprotected model}
\label{activation}
\begin{figure}[htp]
    \centering
    {\includegraphics[width=\columnwidth]{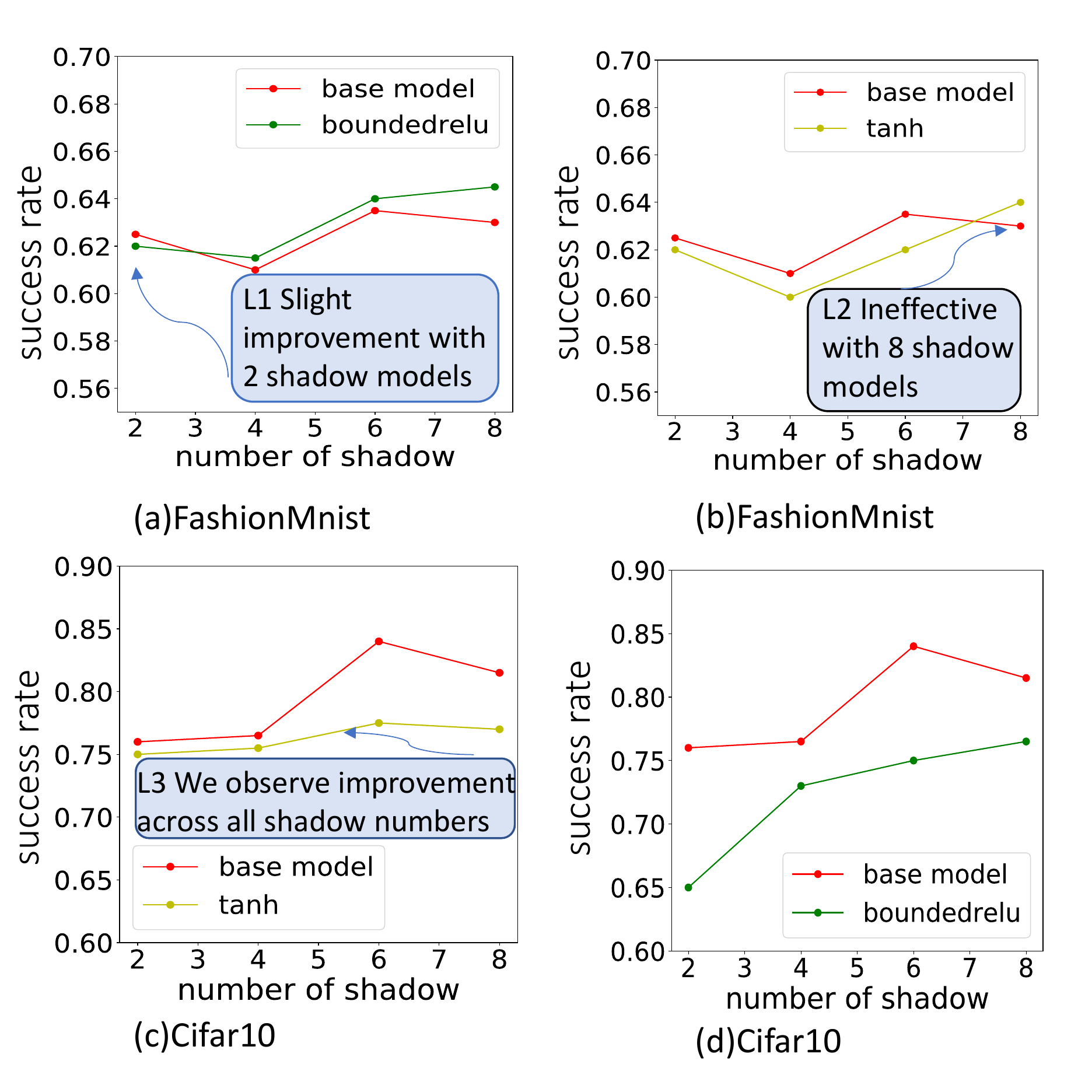}}
    \caption{Impact of varying activation functions on attack success rate of membership inferences attack for an unprotected model.}
    \label{g}
\end{figure} 

Figure \ref{g} illustrates the impact of varying activation functions in the case of unprotected models on the attack success rate of membership inference attack.
To assess the impact of activation functions on the attack success rate, we conducted our experiments on the models mentioned in Table \ref{corr} for FashionMnist and Table \ref{correlations} for CIFAR-10. We replace RELU with BoundedRELU and tanh and then compare attack success rate by varying each time in the number of shadow models. We notice improvement for those activation functions in robustness to the attack for the large model (See L2) compared to the baseline model with RELU while the simple model shows no improvement in robustness with respect to those activation functions (See L1). The significant improvement is in the case of BoundedRELU with CIFAR-10 which reduced the attack success rate by almost 10\% for a $number of shadow = 2$.

\subsection{Impact of normalization on attack success rate for an unprotected model}
\begin{figure}[htp]
    \centering
    {\includegraphics[width=\columnwidth]{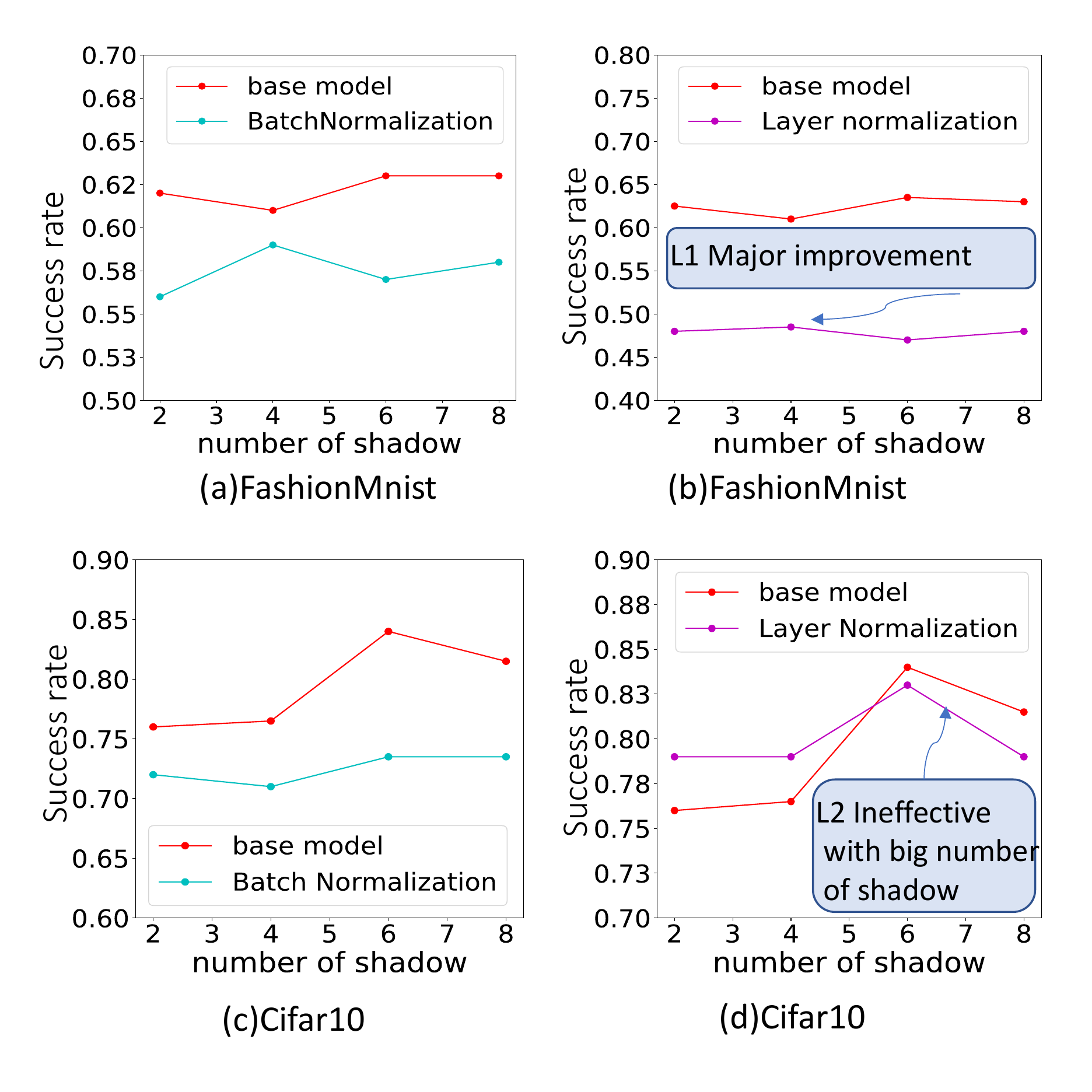}}
    \caption{Impact of normalization layer on attack success rate of membership inferences attack for an unprotected model.}
    \label{hf}
\end{figure} 

Except for the slight drop in robustness for a number of shadows equal to 6 and 8 in Figure \ref{hf} compared to the baseline model due to the experimental bias. All the results show a gain in robustness in the case of batch and layer normalization compared to the baseline model, which was expected since work in \cite{yy} explains the connection between overfitting and membership inference attack effectiveness, the bigger the gap between training and test accuracy the more the attack model can identify which data is used in the training set, and since the normalization technique help reduce overfitting as mentioned in \cite{san} so this is why these techniques mitigate the effect of membership inference attack.
\subsection{Impact of varying learning rate on attack success rate for an unprotected model}
\begin{figure}[htp]
    \centering
    {\includegraphics[width=\columnwidth]{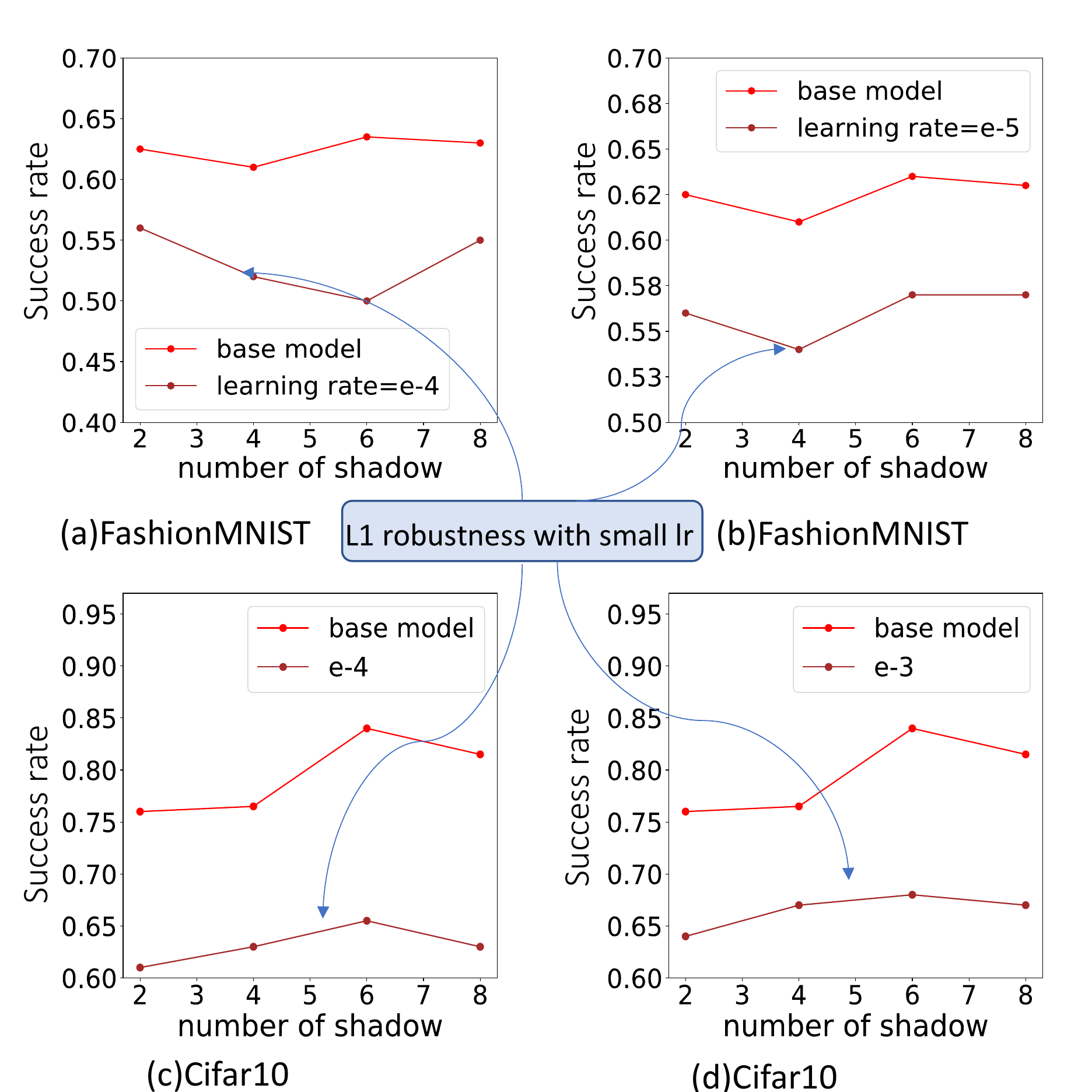}}
    \caption{Impact of varying learning rate on attack success rate of membership inferences attack for an unprotected model.}
    \label{k}
\end{figure} 
Figure \ref{k} shows the impact of learning rate on attack success rate. We notice an improvement in target model robustness against membership inference attack with smaller learning rate compared to baseline model which is trained with a learning rate=$e^-2$, decreasing learning rate diminish roughly attack success rate. This is may be due to the fact that when learning rate is too large often moves too far in the correct direction resulting in an instability of the convergence of our model which causes poor generalization accuracy as mentioned in \cite{wi} while a small learning rate improves generalization accuracy due to the limitation of those overcorrection with a large learning rate.
\subsection{Impact of dropout on attack success rate for an unprotected model}
\begin{figure}[!h]
    \centering
    {\includegraphics[width=\columnwidth]{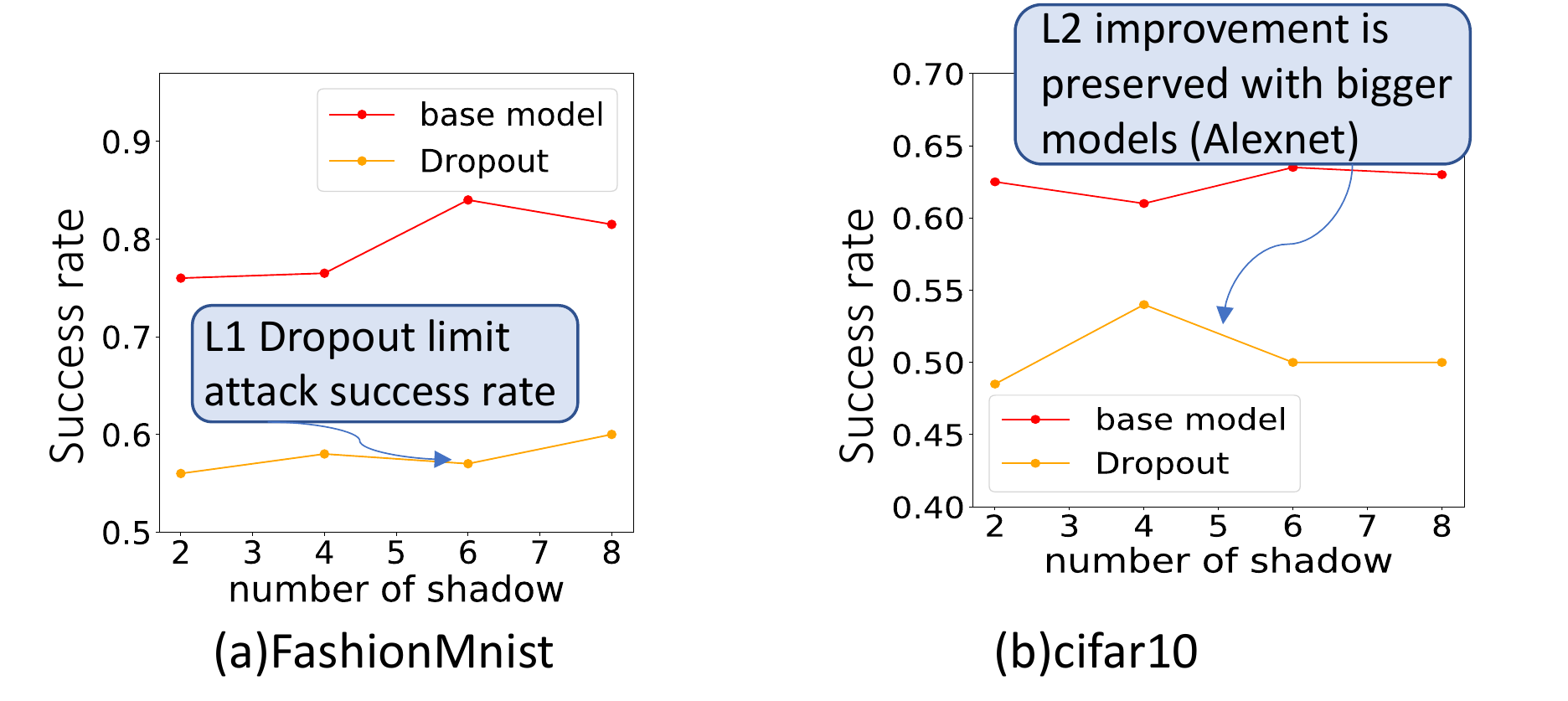}}
    \caption{Evaluating  Impact of dropout on attack success rate for an unprotected model.}
    \label{fig:three graphs}
\end{figure} 
Figure \ref{fig:three graphs} shows the impact of dropout on attack success rate.
As mentioned previously with normalization layer and their effect on limiting overfitting, applying dropout in the training phase help also limiting overfitting which reduces the effectiveness of membership inference attack this is why we see this major improvement for dropout technique especially in the case of CIFAR-10 with a gain of $20\%$ in robustness against the attack.
\begin{figure}[htp]
    \centering
    {\includegraphics[width=\columnwidth]{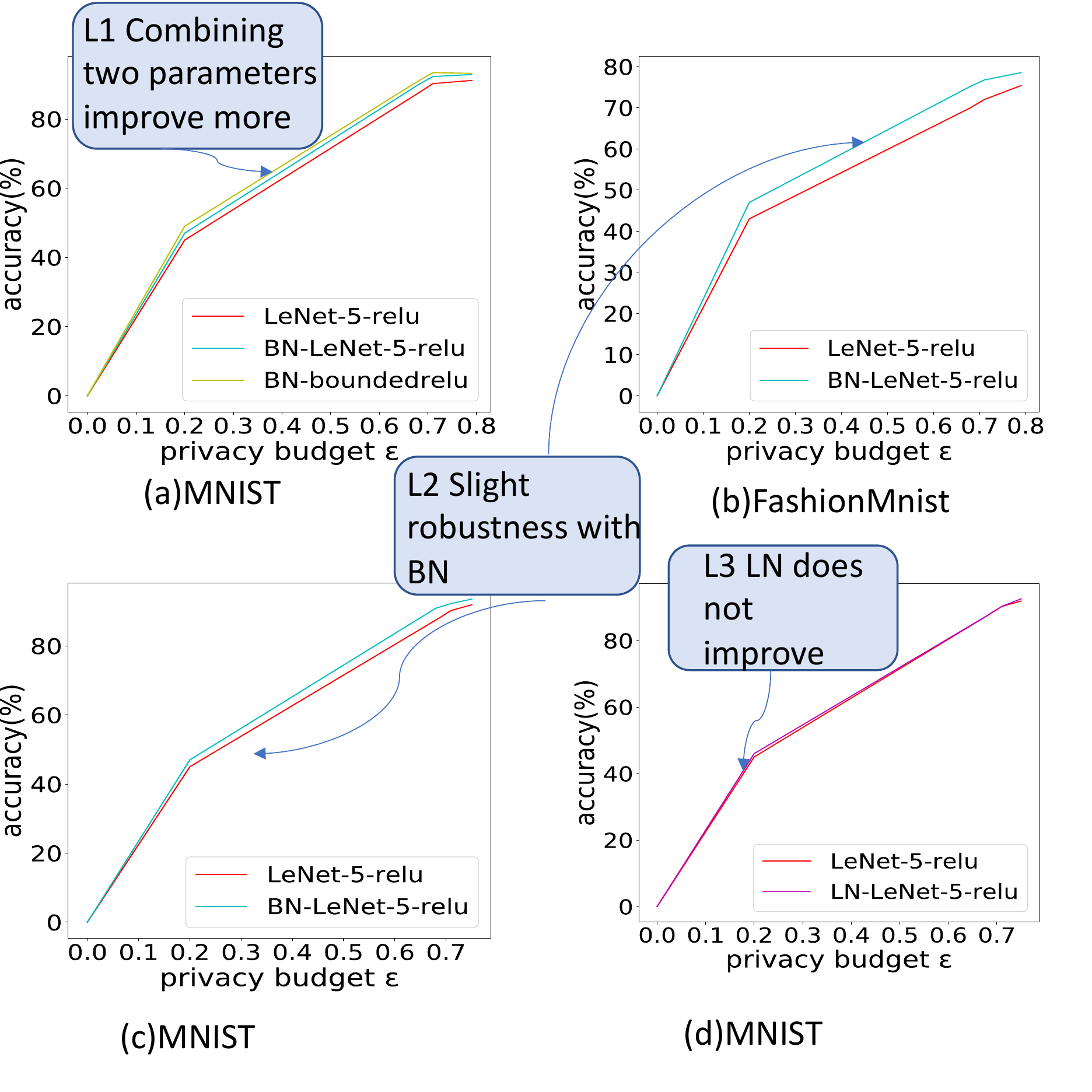}}
    \caption{Impact of normalization layer on privacy vs utility trade-off for a protected model with DPSGD.}
    \label{figH}
\end{figure} 
\subsection{Impact of normalization layer on privacy vs utility trade-off for a protected model}

As Figure \ref{figH} shows Batch normalization improves the utility for the same level of privacy (See L2 in Figure \ref{figH}(b)) due to the fact of stabilizing the learning process and reducing the training time. We also see in Figure \ref{figH} (a) that combining BatchNormalization with BoundedRELU improves the overall trade-off (See L1). For Layer Normalization, we notice no improvement for all privacy levels.
\subsection{Impact of varying activation functions on privacy vs utility trade-off   for a protected model}
\begin{figure}[htp]
    \centering
    {\includegraphics[width=\columnwidth]{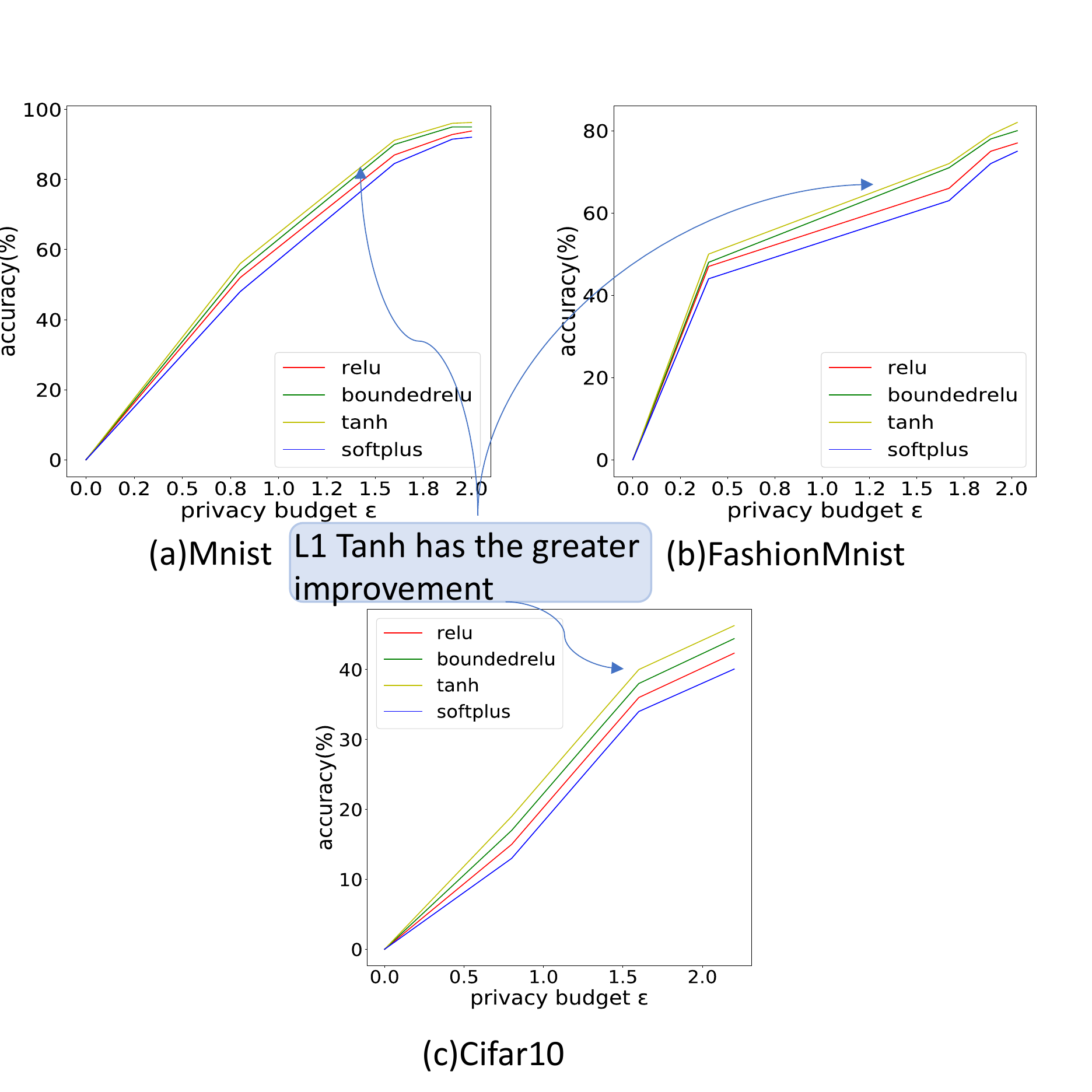}}
    \caption{Accuracy vs privacy budget for different level of noise. }
    \label{figFH}
\end{figure} 

\begin{figure}[htp]
    \centering
    {\includegraphics[width=\columnwidth]{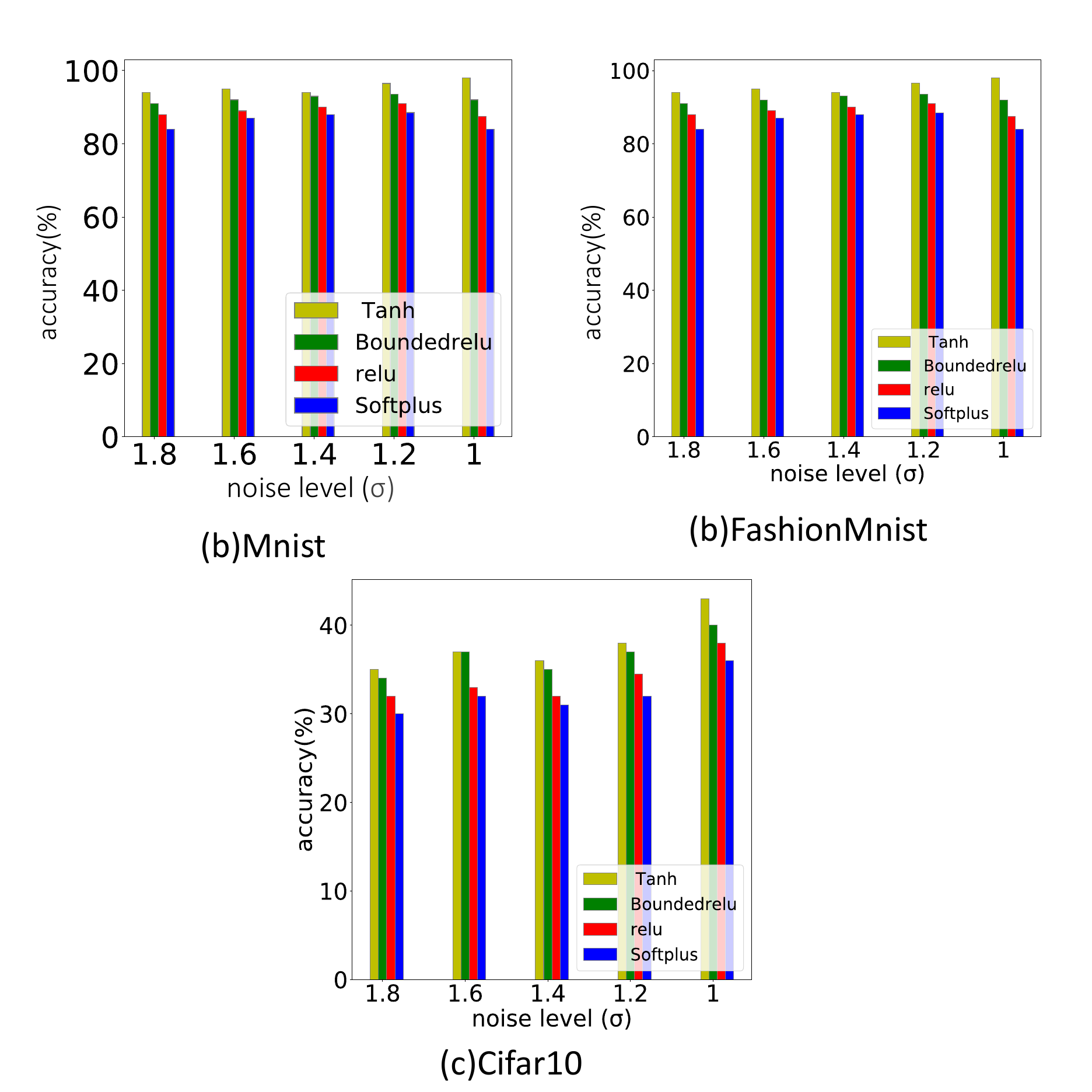}}
    \caption{Average accuracy over the ran epochs for different level of noise.}
    \label{figFHj}
\end{figure} 

The bar chart \ref{figFHj} includes the average accuracy at each noise level for 10 epochs, the different noise levels values are $list=[1.8,1.6,1.4,1.2,1,0.8] $. The learning rate is fixed to $0.1$, $delta=1e-5 $ and $max-per-sample-grad-norm=1.0$ and we set the value $a=2$ which is the parameter of BoundedRELU.

For Figure \ref{figFH}, we trained our model at each noise level for 4 epochs the last noise level is trained for 10 epochs, the different noise levels values are $list_2=[1.4,1.2,1,0.8] $  and each drawn point is the privacy budget and accuracy of the last epoch. The learning rate is fixed to $0.1$, $delta=1e-5$, and $max-per-sample-grad-norm=1.0$ and we set $a=4$ which is the parameter of BoundedRELU.
As Figure \ref{figFHj} and  \ref{figFH} show, bounded activation function outperform unbounded activation functions and this is due to the fact of reducing the impact of clipping gradient and the information discarded at this step which lead to more utility for the same level of privacy. The best performance as the figures show is for Tanh (See L1 in Figure \ref{figFH}) which led to new state-of-the-art results.
\section{Why do bounded activation functions help privacy? }
To respond to this question after the major improvement by bounded activation function seen in section \ref{activation} more specifically in the case of protected model with DPSGD, we use as an example of bounded activation function BoundedRELU which is a simple variant of RELU that differs from it just by putting a threshold a so that if the input exceeds this value it is replaced by a. So its expression is as follows:
$$
f(x)= \begin{cases}x & \text { if } a>x>0 \\a  & \text { if } x>a\\ 0 & \text { otherwise }\end{cases}
$$
\subsection{Impact of activation functions on the gradient magnitude}
 To highlight the impact of BoundedRELU on the gradient size we perform an analytical comparison between RELU and BoundedRELU.
\begin{enumerate}
    \item gradient size with RELU : $$
\frac{\partial L}{\partial w_i}=y_i \cdot \gamma_{i+1}
$$
$\gamma_i=\left\{\begin{array}{c}\left(y_i-\text { Label }\right) \cdot R E L U^{\prime}\left(x_i\right), \text { if i }  \text {the last } \\ \left(\sum \gamma_{i+1} \cdot w_i\right) \cdot R E L U^{\prime}\left(x_i\right), \text { else }\end{array}\right.$
where $x_i$,$y_i$ and $w_i$ are respectively the input, the output and the weights of the i-th layer.
\par To highlight the difference between RELU and BoundedRELU in terms of gradient size we consider the case where $x_i>a$ so$\cdot R E L U^{\prime}\left(x_i\right)=1$ and consequently due to this unconstrained value of the partial derivative the value of the gradient will explode in this case to $
\frac{\partial L}{\partial w_i}=y_i \cdot \gamma_{i+1}
$
\item Gradient size with BoundedRELU : 
$$
BRELU^{\prime}(x)= \begin{cases}1 & \text { if } a>x>0 \\0 & \text { if } x>a\\ 0 & \text { otherwise }\end{cases}
$$
$$
\frac{\partial L}{\partial w_i}=y_i \cdot \gamma_{i+1}
$$
$\gamma_i=\left\{\begin{array}{c}\left(y_i-\text { Label }\right) \cdot BRELU^{\prime}\left(x_i\right), \text { if i the last } \\ \left(\sum \gamma_{i+1} \cdot w_i\right) \cdot BRELU^{\prime}\left(x_i\right), \text { else }\end{array}\right.$
\\Taking the case where $x_i>a$,$\cdot BRELU^{\prime}\left(x_i\right)=0$ so the value of the gradient will be : $
\frac{\partial L}{\partial w_i}=y_i \cdot \gamma_{i+1}=0
$.
We can notice the importance of BoundedRELU compared to RELU in the case where $x_i>a$ in constraining the value of the gradient compared to RELU which stays intact  since its partial derivative is 1 which will have no impact on constraining the derivative in the chain rule.
\end{enumerate}

\begin{figure}[htp]
    \centering
    {\includegraphics[width=0.8\columnwidth]{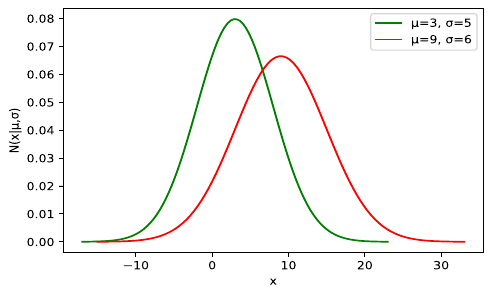}}
    \caption{Norm2 of conv output for CIFAR-10.}
    \label{fhf}
\end{figure} 
Figure \ref{fhf} shows a comparison between the norm2 of the output of the first RELU conv block vs the norm2 of the output of the first BoundedRELU conv block to show the impact of BoundedRELU in our LeNet model for the FashionMnist dataset we can see that the mean value for RELU is higher than the mean value for BoundedRELU which implies the restriction of clipping gradient in DPSGD.
\subsection{DPSGD with BoundedRELU}
\begin{algorithm}
\caption{DP-SGD with BoundedRELU }\label{alg:two}
Input : Training dataset x1,..., xN, Loss function L, Learning
rate $\mu_t$, Noise scale $\sigma$, Batch size b, Parameter of RELU a and clipping threshold C.\\
Initialize $\theta_0$  with uniform initialization;\\
 $a =TuneBound(L, \theta,a)$\;

\For{$t = 0$ to $T-1$}{
  \For{each i in $S_b$}{ $g_r\left(x_l\right) \leftarrow \nabla_{\theta_t} L\left(\theta_l, x_i\right)$\\$
                \bar{g}_t\left(x_i\right) \leftarrow g_r\left(x_i\right)+G\left(0,\sigma^2 C^2\right)
                $\\ }
                $
                \theta_{t+1} \leftarrow \theta_t-\eta_t \bar{g}_t
                $
}

\end{algorithm}
\par To ensure that BoundedRELU works within its optimal performance we need to understand how to improve it, and as mentioned previously its role within DPSGD is to constrain the number of clipping gradients to ensure the minimum information loss. In our case, we clipped the gradient if its value exceed a certain threshold C given that $x_i <a$ so we need to make sure that the chosen value of a minimize the number of values that its gradient exceeds a certain threshold C given that $x_i <a$, after the bound is tuned DPSGD then is executed as always without any changes with its steps. By doing so we soften the effect of clipping gradient in DPSGD and therefore we ensure maximum gain in accuracy for the same level of privacy compared to unbounded RELU.

After trying to evaluate its impact we aimed to integrate BoundedRELU in state-of-the-art works.
\begin{table}
\centering
\begin{tabular}{ |c|c|c|c| } 
\hline
Model for MNIST & Acc & $\epsilon$ \\
\hline
Genetic algorithm with RELU & 73.745 & 1.75\\ 
\textbf{Genetic algorithm with BoundedRELU (ours)}& \textbf{82.99} & 1.74 \\ 
\hline
\end{tabular}
\caption{Comparison of genetic algorithm convergence with RELU vs BoudedRELU for MNIST.}
\label{d1}
\end{table}
Table \ref{d1} and Table \ref{d2} show the results for the genetic optimization algorithm explained in section 4, to run this optimization algorithm we put population strength=10, mutation rate=0.25, the segments=10 and number-of-generation=10, where the search space is $(lr, a)$ and a is the BoundedRELU parameter. The results for RELU are taken from the paper \cite{priyanshu2021efficient} and we repeated the experiments and we got the same results to ensure the fairness of the comparison.
\begin{table}[H]
\centering
\begin{tabular}{ |c|c|c|c| } 
\hline
 Model for CIFAR-10 & Acc & $\epsilon$ \\
\hline
 Genetic algorithm with RELU & 37.999 &  0.599\\ 
\textbf{Genetic algorithm with BoundedRELU (ours)} & 37.88 & \textbf{0.514} \\ 

\hline
\end{tabular}

\caption{Comparison of genetic algorithm convergence with RELU vs BoudedRELU for CIFAR-10. }
\label{d2}
\end{table}
As Table \ref{d3} shows, BoundedRELU outperforms RELU for all three datasets at the same level of privacy, the best improvement is for FashionMnist datasets and by including CIFAR-10 in our experiments we can ensure that our approach works well for more complex datasets. So we can conclude that our new proposed activation function BoundedRELU works well in privacy and is ready to be incorporated in differentially private models rather than unbounded RELU.

Table \ref{d3} presents a summary of some results to demonstrate the effectiveness of BoundedRELU in terms of convergence for each model and also shows a comparison with state-of-the-art.

\begin{table}[H]
\centering
\begin{tabular}{ |c|c|c|c| } 
\hline
 Datasets & Approach & Acc & $\epsilon$ \\
\hline
MNIST     & DP-SGD with RELU & 92.92 &  1.43\\ 
MNIST     & \textbf{DP-SGD with BoundedRELU (ours)} & \textbf{96.02}& 1.43\\ 
CIFAR-10  & DP-SGD WITH RELU & 42.34 &  2.20\\ 
CIFAR-10   &\textbf{DP-SGD with BoundedRELU (ours)} & \textbf{44.42} &  2.20\\ 
FashinMnist & DP-SGD with RELU & 81.9&  2.7\\ 
FashionMnist & \textbf{DP-SGD with BoundedRELU (ours)}& \textbf{84.76 }&  2.7\\ 
\hline
\end{tabular}
\caption{Results summary. }
\label{d3}
\end{table}
\section{Discussion and Concluding Remarks}
In this paper, we present a framework analysis for investigating how the hyperparameter space of the model affects the privacy vs. utility trade-off. Our investigation started with an unprotected model, and we used a membership inference attack to assess privacy leakage. We discovered that although some parameters support robustness, others do not. Except in the case of a large model (Alexnet) with a large number of shadows, normalization techniques (Batch and layer normalization) represent the main improvement. We discovered that dropout helps in all numbers of shadows and lessens the impact of membership inference attacks. Additionally, we discovered that small learning rates enhance model privacy more so than big learning rates. Finally, we address the privacy implications of several activation functions in this case.
The impact of hyperparameter space on the privacy vs. utility trade-off in the protected model was then assessed using DPSGD.
In this instance, we discovered that bounded activation functions significantly enhanced the model's output and produced new state-of-the-art results compared to what is commonly used. We highlighted batch normalization's effects and how they affect the model, as a result, it is clear that this parameter helps while layer normalization does not. We can also see how several parameters might be combined to increase model privacy.
Finally, we provide an analysis in the case of a bounded activation function and use the example of BoundedRELU to study this case.

Since the combination of two parameters shows an improvement in our studies, one can propose an optimization method, such as a genetic algorithm, or perform a grid search optimization on the provided set of parameters to build a more efficient model.
A different direction is to attempt to incorporate those parameters to produce new, effective privacy-preserving designs.
Neural architecture search (NAS) is another topic that requires investigation, one can leverage this work to build more efficient architectures.


\nocite{*}
\bibliographystyle{IEEEtran}
\bibliography{bib}

\end{document}